\documentclass[10pt,twocolumn,letterpaper]{article}

\usepackage{cvpr}              

\usepackage{graphicx}
\usepackage{amsmath}
\usepackage{amssymb}
\usepackage{booktabs}

\usepackage{multirow} 
\usepackage{array}    

\usepackage{caption}
\usepackage{enumitem}
\setitemize{noitemsep,topsep=0pt,parsep=0pt,partopsep=0pt}
\usepackage[pagebackref,breaklinks,colorlinks]{hyperref}

\usepackage[capitalize]{cleveref}
\crefname{section}{Sec.}{Secs.}
\Crefname{section}{Section}{Sections}
\Crefname{table}{Table}{Tables}
\crefname{table}{Tab.}{Tabs.}

\def\cvprPaperID{2237} 
\def\confName{CVPR}
\def\confYear{2023}

\begin{document}

\title{SMAE: Few-shot Learning for HDR Deghosting with Saturation-Aware \\Masked Autoencoders}

\author{Qingsen Yan$^\dag$$^1$ \quad Song Zhang$^\dag$$^2$ \quad Weiye Chen$^\dag$$^2$ \quad Hao Tang$^3$ \quad Yu Zhu$^1$ \\ Jinqiu Sun$^1$ \quad Luc Van Gool$^3$ \quad Yanning Zhang$^1$\thanks{$\dag$~The first three authors contributed equally to this work.
This work was partially supported by NSFC (U19B2037, 61901384), Natural Science Basic Research Program of Shaanxi (2021JCW-03, 2023-JC-QN-0685), and National Engineering Laboratory for Integrated Aero-Space-Ground-Ocean Big Data Application Technology. Corresponding author: Yu Zhu.}\\
$^1$Northwestern Polytechnical University \quad
$^2$Xidian University \quad
$^3$CVL, ETH Zurich
}



\maketitle

\begin{abstract}
Generating a high-quality High Dynamic Range (HDR) image from dynamic scenes has recently been extensively studied by exploiting Deep Neural Networks (DNNs). 
Most DNNs-based methods require a large amount of training data with ground truth, requiring tedious and time-consuming work.  
Few-shot HDR imaging aims to generate satisfactory images with limited data. However, it is difficult for modern DNNs to avoid overfitting when trained on only a few images.
In this work, we propose a novel semi-supervised approach to realize few-shot HDR imaging via two stages of training, called SSHDR. Unlikely previous methods, directly recovering content and removing ghosts simultaneously, which is hard to achieve optimum, we first generate content of saturated regions with a self-supervised mechanism and then address ghosts via an iterative semi-supervised learning framework.
Concretely, considering that saturated regions can be regarded as masking Low Dynamic Range (LDR) input regions, we design a Saturated Mask AutoEncoder (SMAE) to learn a robust feature representation and reconstruct a non-saturated HDR image.
We also propose an adaptive pseudo-label selection strategy to pick high-quality HDR pseudo-labels in the second stage to avoid the effect of mislabeled samples.
Experiments demonstrate that SSHDR outperforms state-of-the-art methods quantitatively and qualitatively within and across different datasets, achieving appealing HDR visualization with few labeled samples.
\end{abstract}

\section{Introduction}
\label{sec:intro}
Standard digital photography sensors are unable to capture the wide range of illumination present in natural scenes, resulting in Low Dynamic Range (LDR) images that often suffer from over or underexposed regions, which can damage the details of the scene. High Dynamic Range (HDR) imaging has been developed to address these limitations. This technique combines several LDR images with different exposures to generate an HDR image. While HDR imaging can effectively recover details in static scenes, it may produce ghosting artifacts when used with dynamic scenes or hand-held camera scenarios.

\begin{figure}[t!]
    \centering
    \includegraphics[width=1\linewidth]{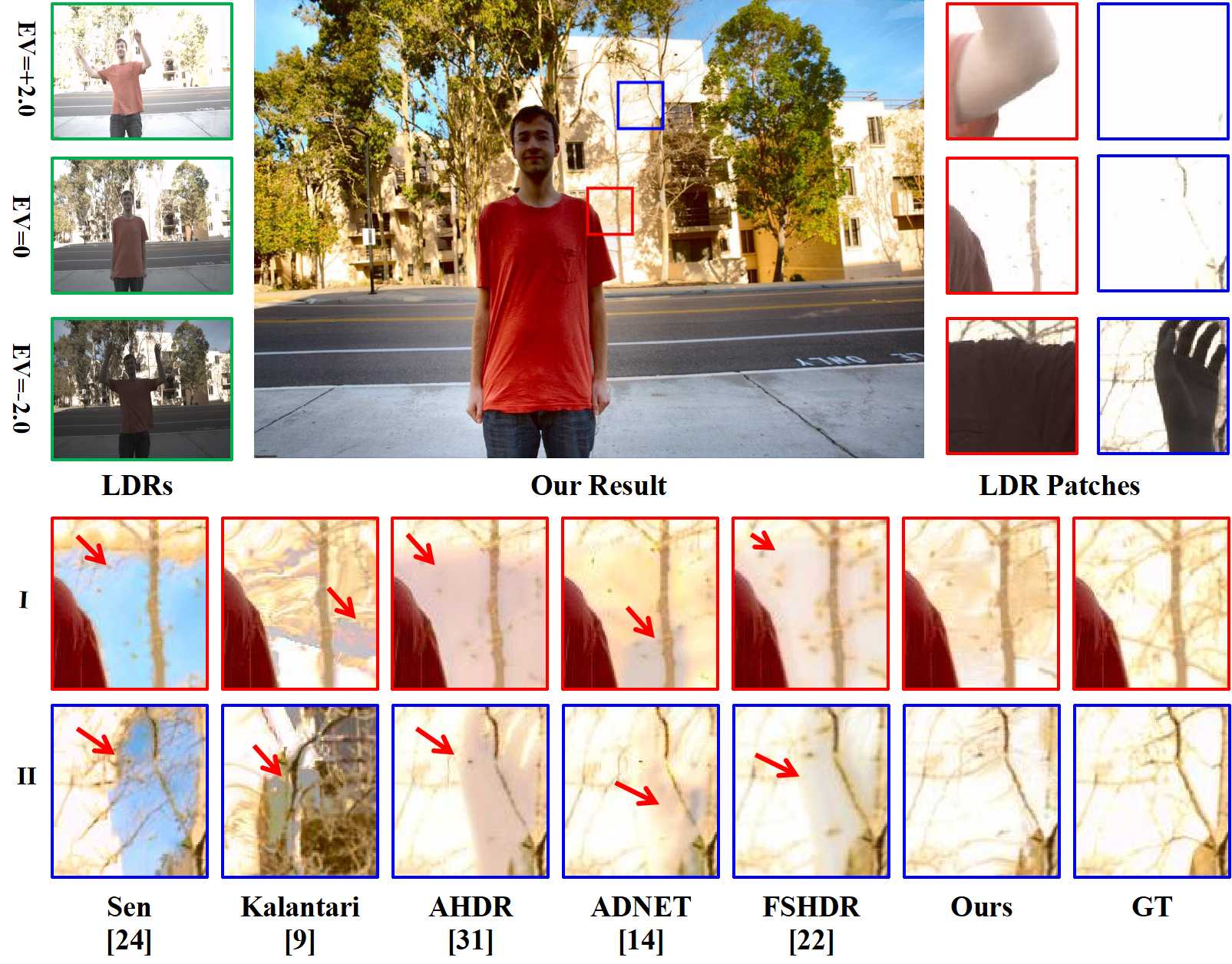}
    \caption{The proposed method generates high-quality images with few labeled samples when compared with several methods.}
    \label{fig:intro}
    \vspace{-0.4cm}
\end{figure}

Historically, various techniques have been suggested to address such issues, such as alignment-based methods \cite{Bogoni2000Extending,KangSig2003High,Tomaszewska2007Image,yan2019robust}, patch-based methods \cite{Sen2012Robust, Hu2013HDR, Ma2017Robust}, and rejection-based methods \cite{Grosch2006Fast, Pece2010Bitmap, Zhang2011Gradient, Lee2014Ghost, Yan2017High,Oh2015Robust}. Two categories of alignment-based approaches exist: rigid alignment (\eg, homographies) that fail to address foreground motions, and non-rigid alignment (\eg, optical flow) that is error-prone.
Patch-based techniques merge similar regions using patch-level alignment and produce superior results, but suffer from high complexity. Rejection-based methods aim to eliminate misaligned areas before fusing images, but may result in a loss of information in motion regions.

As Deep Neural Networks (DNNs) become increasingly prevalent, the DNN-based HDR deghosting methods \cite{Kalantari2017Deep,yan2022high,yan2020ghost} achieve better visual results compared to traditional methods. 
However, these alignment approaches are error-prone and inevitably cause ghosting artifacts (see Figure \ref{fig:intro} Kalantari's results). AHDR \cite{Yan2019attention,yan2022dual} proposes spatial attention to suppress motion and saturation, which effectively alleviate misalignment problems. Based on AHDR, ADNET \cite{Liu2021ADNet} proposes a dual branch architecture using spatial attention and PCD-alignment \cite{wang2019edvr} to remove ghosting artifacts. All these above methods directly learn the complicated HDR mapping function with abundant HDR ground truth data. However, it’s challenging to collect a large amount of HDR-labeled data. The reasons can be attributed to that 1) generating a ghost-free HDR ground truth sample requires an absolute static background, and 2) it is time-consuming and requires considerable manpower to do manual post-examination. This generates a new setting that only uses a few labeled data for HDR imaging.

Recently, FSHDR \cite{prabhakar2021labeled} attempts to generate a ghost-free HDR image with only few labeled data.
They use a preliminary model trained with a large amount of unlabeled dynamic samples, and a few dynamic and static labeled samples to generate HDR pseudo-labels and synthesize artificial dynamic LDR inputs to improve the model performance of dynamic scenes further.
This approach expects the model to handle both the saturation and the ghosting problems simultaneously, but it is hard to achieve under the condition of few labeled data, especially misaligned regions caused by saturation and motion (see Figure \ref{fig:intro} FSHDR).
In addition, FSHDR uses optical flow to forcibly synthesize dynamic LDR inputs from poorly generated HDR pseudo-labels, the errors in optical flow further intensify the degraded quality of artificial dynamic LDR images, resulting in an apparent distribution shift between LDR training and testing data, which hampers the performance of the network.

The above analysis makes it very challenging to directly generate a high-quality and ghost-free HDR image with few labeled samples. A reasonable way is to address the saturation problems first and then cope with the ghosting problems with a few labeled samples. In this paper, we propose a semi-supervised approach for HDR deghosting, named SSHDR, which consists of two stages: self-supervised learning network for content completion and sample-quality-based iterative semi-supervised learning for deghosting.
In the first stage, we pretrain a Saturated Mask AutoEncoder (SMAE), which learns the representation of HDR features to generate content of saturated regions by self-supervised learning.
Specifically, considering that the saturated regions can be regarded as masking the short LDR input patches, inspired by \cite{he2022masked}, we randomly mask a high proportion of the short LDR input and expect the model to reconstruct a no-saturated HDR image from the remaining LDR patches in the first stage.
This self-supervised approach allows the model to recover the saturated regions with the capability to effectively learn a robust representation for the HDR domain and map an LDR image to an HDR image. 
In the second stage, to prevent the overfitting problem with a few labeled training samples and make full use of the unlabeled samples, we iteratively train the model with a few labeled samples and a large amount of HDR pseudo-labels from unlabeled data.
Based on the pretrained SMAE, a sample-quality-based iterative semi-supervised learning framework is proposed to address the ghosting artifacts.
Considering the quality of pseudo-labels is uneven, we develop an adaptive pseudo-labels selection strategy to pick high-quality HDR pseudo-labels (\ie, well-exposed, ghost-free) to avoid awful pseudo-labels hampering the optimization process. This selection strategy is guided by a few labeled samples and enhances the diversity of training samples in each epoch.
The experiments demonstrate that our proposed approach can generate high-quality HDR images with few labeled samples and achieves state-of-the-art performance on individual and cross-public datasets. 
Our contributions can be summarized as follows:
\begin{itemize}[leftmargin=*]
\item We propose a novel and generalized HDR self-supervised pretraining model, which uses mask strategy to reconstruct an HDR image and addresses saturated problems from one LDR image. 
\item We propose a sample-quality-based semi-supervised training approach to select well-exposed and ghost-free HDR pseudo-labels, which improves ghost removal.


\item We perform both qualitative and quantitative experiments, which show that our method achieves state-of-the-art results on individual and cross-public datasets.
\end{itemize}

\section{Related Work}
\subsection{HDR Deghosting Methods}
The existing HDR deghosting methods include four categories: alignment-based method, patch-based method, rejection-based method, and CNN-based method.

\noindent\textbf{Alignment-based Method.}
Rigid or non-rigid registration is mainly used in alignment-based approaches. Bogoni \cite{Bogoni2000Extending} estimated flow vectors to align with the reference images. Kang \etal \cite{KangSig2003High} utilized optical flow to align images in the luminance domain to remove ghosting artifacts. Tomaszewska \etal \cite{Tomaszewska2007Image} used SIFT feature to perform global alignments. Since the dense correspondence computed by alignment methods are error-prone, they cannot handle large motion and occlusion.

\noindent\textbf{Rejection-based Method.}
Rejection-based methods detect and eliminate motion from static regions. Then they merge static inputs to get HDR images. Grosch \etal \cite{Grosch2006Fast} estimated a  motion map and used it to generate ghost-free HDR. Zhang \etal \cite{Zhang2011Gradient} obtained a motion weighting map using quality measurement on image gradients. Lee \etal \cite{Lee2014Ghost} and Oh \etal \cite{Oh2015Robust} detected motion regions using rank minimization. However, rejection-based methods remove the misalignment of regions. It will result in a lack of content in moving regions.

\noindent\textbf{Patch-based Method.}
Patch-based methods use patch-level alignment to merge similar contents. Sen \etal \cite{Sen2012Robust} proposed a patch-based energy minimization approach that optimizes alignment and reconstruction simultaneously. Hu \etal \cite{Hu2013HDR} utilized a patch-match mechanism to produce aligned images. Although these methods have good performance, they suffer from high computational costs.

\begin{figure*}[t]
\centering
\includegraphics[width=1\linewidth]{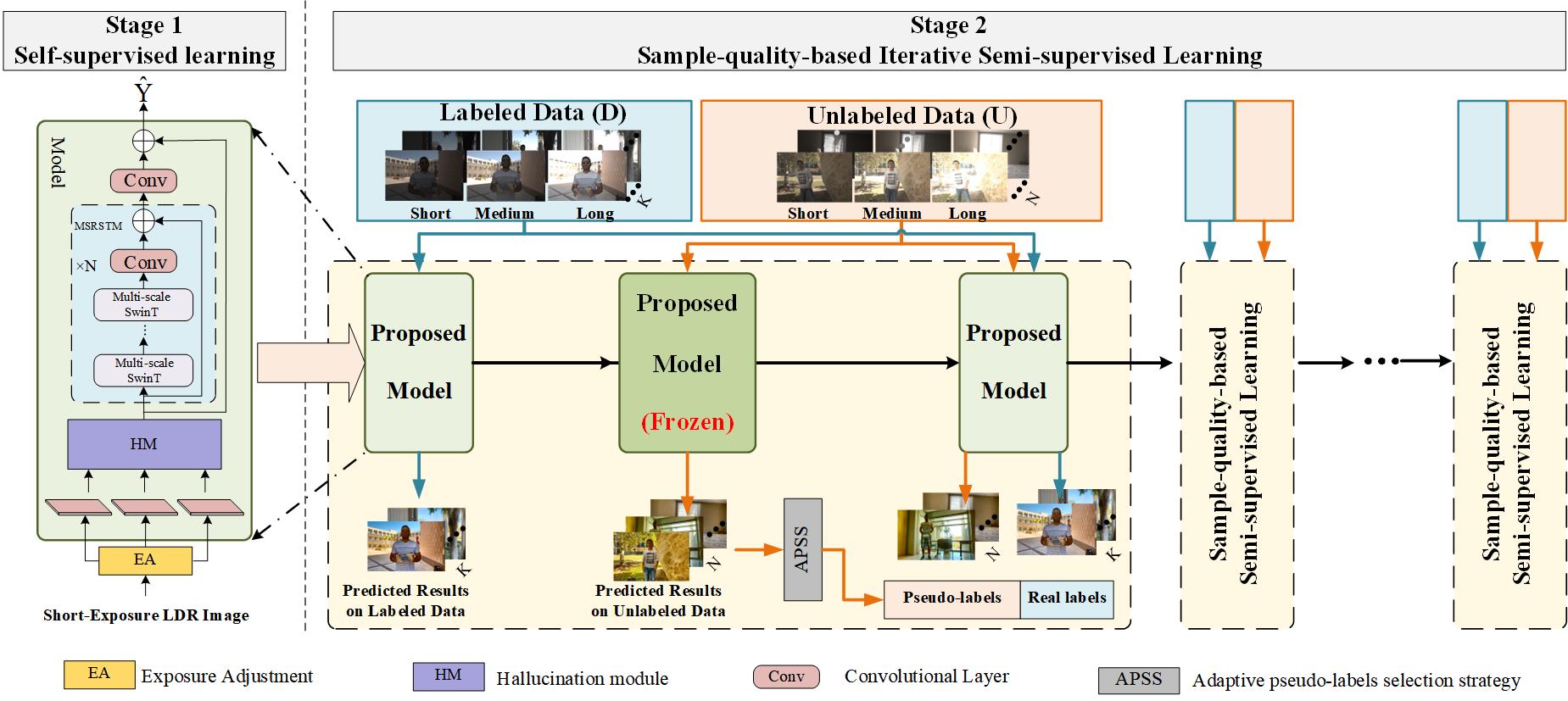}
\caption{The overview of our framework.}
\label{fig:Fig_method}
\vspace{-0.4cm}
\end{figure*}

\noindent\textbf{CNN-based Method.}
Kalantari \etal \cite{Kalantari2017Deep} used a CNN network to fuse LDR images that are aligned with optical flow. Wu \etal \cite{Wu2018Deep} used homography to align the camera motion and reconstructed HDR images by CNN.
Yan \etal \cite{Yan2019attention} proposed an attention mechanism to suppress motion and saturation. Yan \etal \cite{Yan2020Deep} designed a nonlocal block to release the constraint of locality receptive field for global merging HDR. Niu \etal \cite{Niu2021Hdr} proposed HDR-GAN to recover missing content using generative adversarial networks. 
Ye \etal \cite{Ye2021Progressive} proposed multi-step feature fusion to generate ghost-free images. Liu \etal \cite{Liu2021ADNet} utilized the PCD alignment subnetwork to remove ghosts.
However, these methods require a large number of labeled samples, which is difficult to collect.

\subsection{Few-shot Learning (FSL)}
Humans can successfully learn new ideas with relatively little supervision. Inspired by such ability, FSL aims to learn robust representations with few labeled samples. There are three main categories for FSL methods: data-based category \cite{benaim2018one,qi2018low,shyam2017attentive}, which augment the experience with prior knowledge; model-based category \cite{miller2016key,sukhbaatar2015end,bertinetto2016learning}, which shrinks the size of the hypothesis space using prior knowledge; algorithm-based category \cite{finn2017model,lee2018gradient,yoon2018bayesian}, which modifies the search for the optimal hypothesis using prior knowledge. For HDR deghosting, Prabhakar \etal \cite{prabhakar2021labeled} proposed a data-based category deghosting method, which uses artificial dynamic sequences synthesis for motion transfer. Still, it is hard to handle both the saturation and the ghosting problems simultaneously with few labeled data.

\section{The Proposed Method}
\subsection{Data Distribution}
Following the setting of few-shot HDR imaging \cite{prabhakar2021labeled}, we utilize 1) $N$ dynamic unlabeled LDR samples $U{=}\{L^U_1,\dots,L^U_N\}$, where each $L^U$ consists of three LDRs ($X^U_1$,$X^U_2$,$X^U_3$) with different exposures. 2) $M$ static labeled LDR samples $S{=}\{L^S_1,\dots,L^S_M\}$, where each $L^S$ consists of three LDRs ($X^S_1$,$X^S_2$,$X^S_3$) with different exposures and ground truth $Y^S$. 3) $K$ dynamic labeled LDR samples $D{=}\{L^D_1,\dots,L^D_K\}$, where each $L^D$ consists of three LDRs ($X^D_1$,$X^D_2$,$X^D_3$) and ground truth $Y^D$. Since it is difficult to collect labeled samples, we set $K$ to be less than or equal to 5, and $M$ is fixed at 5. While it is easy to capture unlabeled samples, $N$ can be arbitrary.

\subsection{Model Overview}
Generating a non-saturated and ghost-free HDR image with few labeled samples is challenging. It is a proper way to address saturated problems first and then handle ghosting problems.
As shown in Figure \ref{fig:Fig_method}, we propose a semi-supervised approach for HDR deghosting. Our approach consists of two stages: a self-supervised learning network for content completion and a sample-quality-based iterative semi-supervised learning for deghosting.
In the first stage, we propose a multi-scale Transformer model based on self-supervised learning with a saturated-masked autoencoder to make it capable of recovering saturated regions. In a word, we randomly mask LDR patches and reconstruct non-saturated HDR images from the remaining LDR patches.

In the second stage, we propose a sample-quality-based iterative semi-supervised learning approach that learns to address ghosting problems. We finetune the pretrained model based on the first stage with a few labeled samples. Then, we iteratively train the model with labeled samples and unlabeled samples with pseudo-labels. Considering that the HDR pseudo-labels inevitably contain saturated and ghosting regions, which deteriorate the model performance, we propose an adaptive pseudo-labels selection strategy to pick high-quality HDR pseudo-labels to avoid awful pseudo-labels hampering the optimization process.

\subsection{Self-supervised Learning Stage}
\begin{figure}[t!]
    \centering
    \includegraphics[width=1\linewidth]{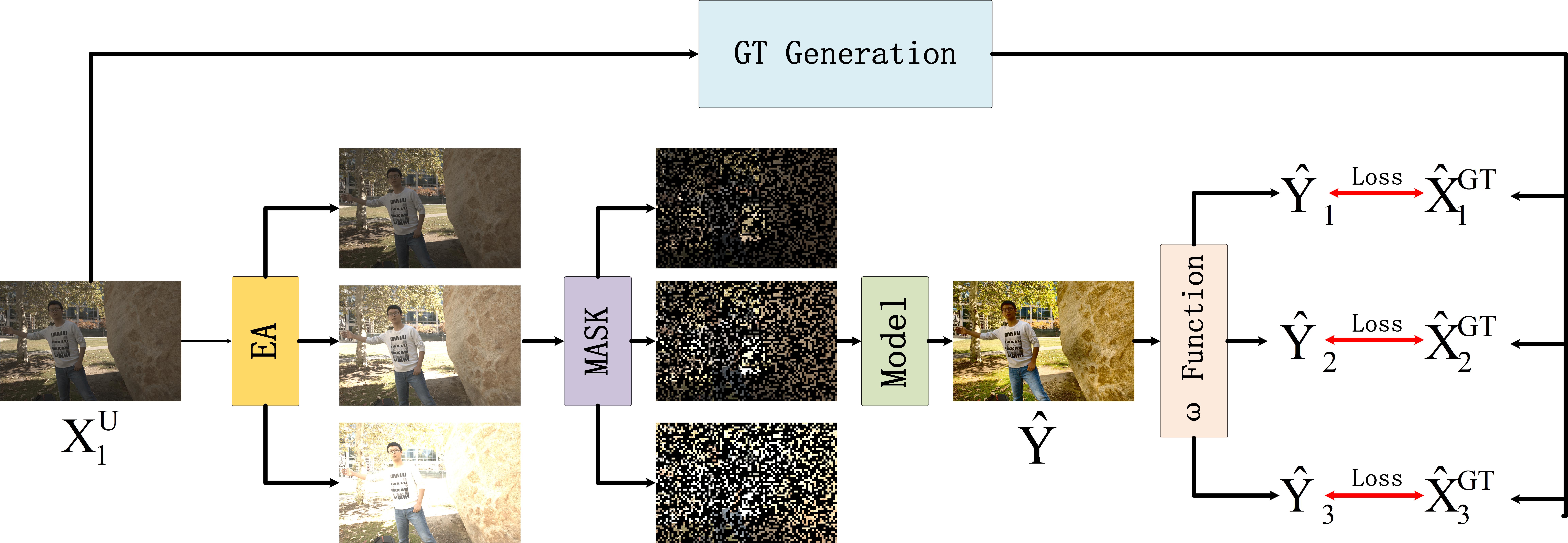}
    \caption{The detailed procedure of Stage 1. To recover the saturated regions, we utilize the short exposure frame as input and ground truth.}
    \label{fig:same}
    \vspace{-0.4cm}
\end{figure}

\noindent\textbf{Input.} Considering that there are more saturated regions in the medium ($X^U_2$) and long exposure frames ($X^U_3$) of unlabeled data $U$, we first transform the short exposure frame ($X^U_1$) into a new medium ($X^U_{2^{’}}$) and long exposure frames ($X^U_{3^{’}}$) by exposure adjustment,
\vspace{-0.1cm}
\begin{equation}
    X^U_{i^{’}} = clip((\frac{{(X^U_1)}^{\gamma} {\times}t_i} {t_1})^\frac{1}{\gamma}), \quad i=2,3.
\end{equation}
Then following previous work\cite{Kalantari2017Deep,Wu2018Deep}, we map the LDR input images $X^U_1,X^U_{2^{’}},X^U_{3^{’}}$ to HDR domain by gamma correction to get $H_i$,
\vspace{-0.1cm}
\begin{equation}
    H_{i^{’}} = {{(X^U_{i^{’}})}^{\gamma}} / {t_i}.
\end{equation}
Note that $X^U_{1^{’}} {=} X^U_1$, $t_i$ denotes the exposure time of LDR image $X_i$, ${\gamma}$ represents the gamma correction parameter, we set ${\gamma}$ to 2.2. Then, we concatenate $X_{i^{’}}$ and $H_{i^{’}}$ along the channel dimension to get a 6-channel input $I_i {=} [X_{i^{’}},H_{i^{’}}]$, we subsequently mask the input $I_i$ patches to get $I^{'}_i$. Concretely, we divide the input into non-overlapping patches and randomly mask a subset of these patches with a high mask ratio (75\%) (see Figure \ref{fig:same}). Note that the patch size is 8{$\times$}8. Considering that the masking strategy is another way to destruct the saturated regions, we intend the model to learn a robust representation to recover these saturated regions. Finally, $I^{'} {=} {\{I^{'}_1,I^{'}_2,I^{'}_3\}}$ is the input of the model.

\noindent\textbf{Model.} Our SMAE self-supervised training-based multi-scale Transformer consists of a feature extraction module, hallucination module, and Multi-Scale Residual Swin Transformer fusion Module (MSRSTM). The details in our model are included in the Appendix.




\textit{Hallucination Module.} We first adopt three convolutional layers to extract shallow feature $F_{i}$. Then, we divide the shallow feature $F_{i}$ into non-overlapping patches $\overline{{F_{i}}}$, and map each patch $\overline{{F_{i}}}$ into query, keys and values. Subsequently, we calculate the similarity map between $\overline{q}$ and $\overline{k}$, and perform the Softmax function to get the attention weight. Finally, we apply the attention weight to $\overline{v}$ to get $F^{i}_s$,
\vspace{-0.1cm}
\begin{equation}
    \begin{aligned}
\overline{q} & =\overline{F_{2}}W_{q}, \quad \overline{k_{i}}=\overline{F_{i}}W_{k}, \quad \overline{v_{i}}=\overline{F_{i}}W_{v}, \quad i=1,3 \\
    F^{i}_s & = {\rm Softmax}(\overline{q} \overline{k_{i}}^{T} / \sqrt{d} + b)\overline{v_{i}},
    \end{aligned}
\end{equation}
where $b$ represents a learnable position encoding, $d$ denotes the dimension of $\overline{q}$.


\textit{MSRSTM.} To merge more information from different exposure regions,  inspired by \cite{liang2021swinir}, we propose a Multi-Scale Residual Swin Transformer Module (MSRSTM). First, $F^{1}_s,F^{2}_s,F^{3}_s$ is concatenated along the channel dimension to get the input of MSRSTM. Note that $F^{2}_s$ denotes $F_2$. Then, MSRSTM merges a long range of information from different exposure regions. MSRSTM consists of multiple multi-scale Swin Transformer layers (STL), a few convolutional layers, and a residual connection. Given the input feature $F^{N-1}_{out,i}$ of $i$-th MSRSTM, the output $F^{N}_{out,i}$ of MSRSTM can be formulated as follows :
\vspace{-0.1cm}
\begin{equation}
\begin{split}
    F^{N}_{STL,i} = Conv((Concat(STL^{N,l_1}_{i}(F^{N-1}_{out,i}),\\STL^{N,l_2}_{i}(F^{N-1}_{out,i}),STL^{N,l_3}_{i}(F^{N-1}_{out,i})),
\end{split}
\end{equation}
\begin{equation}
    F^{N}_{out,i} = Conv(F^{N}_{STL,i}) + F^{N-1}_{out,i},
\end{equation}
where $STL^{N,l_j}_{i}(\cdot)$ represents the $N$-th Swin Transformer layer of the $l_j$ scale in the $i$-th MSRSTM, $F^{N-1}_{out,i}$ denotes the input feature of the $N$-th Swin         Transformer layer in the $i$-th MSRSTM.

\noindent\textbf{Loss Function.} Since unlabeled samples do not have HDR ground truth labels, we calculate the self-supervised loss in the LDR domain. We first use function $\omega$ to transform the predicted HDR image $\hat{Y}$ to short, medium, and long exposure LDR images $\hat{Y_i}$,
\vspace{-0.1cm}
\begin{equation}
    \hat{Y_i} = \omega(\hat{Y}) = (\hat{Y} \times {t_i})^{\frac{1}{\gamma}}.
\end{equation}
To recover the saturated regions, we transform the short exposure frame (since the predicted HDR in this stage is aligned to the short exposure frame) to new short, medium, and long exposure frames by ground truth generation. Then, we regard the new exposure frames as the ground truth $X^{GT}_{i}$ of the model,
\vspace{-0.1cm}
\begin{equation}
    X^{GT}_i = (\frac{{(X^U_1)}^{\gamma} {\times}t_i} {t_1})^{\frac{1}{\gamma}}, \quad i=1,2,3.
\end{equation}
Finally, we calculate $L_1$ self-supervised loss between $\hat{Y}_i$ and $X^{GT}_i$,
\vspace{-0.3cm}
\begin{equation}
    L_{SSL} = \sum_{i=1}^3 ||\hat{Y}_i-X^{GT}_i||_1.
\end{equation}

\subsection{Semi-supervised Learning Stage}
\noindent\textbf{Finetune.} 
At the beginning of this stage, to improve the saturated regions and further learn to handle ghosting regions, we first finetune the pretrained model with a few dynamic samples $D$ and static labeled samples $S$. 
Here we apply $\mu$-law to map the linear domain image to the tonemapped domain image,
\vspace{-0.3cm}
\begin{equation}
    T(x)=\frac{log(1+\mu x)}{log(1+\mu)},
\end{equation}
where $T(x)$ is the tonemap function, $\mu{=}5000$. Then we calculate the reconstruction loss $L_{recon}$ and perceptual loss $L_{percep}$ between the predicted HDR $\hat{Y}^D_0,\hat{Y}^S_0$ and ground truth HDR ${Y}^D_0,{Y}^S_0$,
\vspace{-0.1cm}
\begin{equation}
    L_{recon} = ||T(\hat{Y}^D_0)-T({Y}^D_0)||_1 + ||T(\hat{Y}^S_0)-T({Y}^S_0)||_1,
\end{equation}
\begin{equation}
\begin{split}
    L_{percep} = ||\phi_{i,j}(T(\hat{Y}^D_0))-\phi_{i,j}(T({Y}^D_0))||_1 \\  + ||\phi_{i,j}(T(\hat{Y}^S_0))-\phi_{i,j}(T({Y}^S_0))||_1,
\end{split}
\end{equation}
\begin{equation}
    L_{finetune} = L_{recon} + \lambda L_{percep},
\end{equation}
where $\phi_{i,j}$ represents the $j$-th convolutional layer and the $i$-th max-pooling layer in VGG19, $\lambda{=}1e^{-2}$.

\noindent\textbf{Iteration.} To prevent the overfitting problem with a few labeled training samples and exploit unlabeled samples, we further generate the pseudo-labels $\hat{Y}^U_t$ of unlabeled data. Concretely, we iteratively and adaptively train the model with a few dynamic and static samples $D$ and $S$ and a large number of unlabeled samples $U$. Specifically, at timestep $t$, we use model $N_t$ to predict the pseudo-labels $\hat{Y}^U_t$ of unlabeled data. Then, we train the model $N_t$ with a few labeled and pseudo-labeled samples to get the model $N_{t+1}$ at timestep $t+1$. Note that we use finetune model to generate unlabel HDR pseudo-labels $\hat{Y}^U_0$ at timestep $t{=}0$. Finally, at each timestep in the refinement stage, we calculate the reconstruction loss and perceptual loss as follows,
\vspace{-0.1cm}
\begin{equation}
\begin{split}
    L_{Iteration} = L^D_{recon,t+1} + L^S_{recon,t+1} + \sum_{i=1}^N
    W^{U_i}_{t+1} L^{U_i}_{recon,t+1} \\
    + \lambda (L^D_{percep,t+1} + L^S_{percep,t+1} + \sum_{i=1}^N
    W^{U_i}_{t+1} L^{U_i}_{percep,t+1}),
\end{split}
\end{equation}
where $\lambda{=}1e^{-2}$. $W^{U_i}_{t+1}$ is the weight factor of unlabeled data $U_i$. To get loss weight $W^{U_i}_{t+1}$, please refer to the next section in detail.

\noindent\textbf{APSS.} Since the HDR pseudo-labels inevitably contain saturated and ghosted samples, we propose an Adaptive Pseudo-labels Selection Strategy (APSS) to pick well-exposed and ghost-free HDR pseudo-labels to avoid awful pseudo-labels hampering the optimization process. Specifically, at timestep $t$, we use model $N_t$ to predict HDR images with dynamic and static labeled samples $\hat{Y}^D_t$ and $\hat{Y}^S_t$. Then we use function $\omega$ to map the predicted HDR image to medium exposure image $\tilde{Y}^{D \cup S}_t$ and calculate the loss between $\tilde{Y}^{D \cup S}_t$ and original medium exposure LDR image $X^{D \cup S}_{2,t}$ in well exposure regions to get $L^{D \cup S}_{select,t}$,
\vspace{-0.1cm}
\begin{equation}
\begin{split}
    L^{D \cup S}_{select,t} = ||mask(\omega(\hat{Y}^D_t))-mask(\omega({X}^D_{2,t}))||_1 \\ 
    + ||mask(\omega(\hat{Y}^S_t))-mask(\omega({X}^S_{2,t}))||_1,
\end{split}
\end{equation}
where $mask(\cdot)$ denotes masking the over and under-exposure regions. Subsequently, we sort all patches' losses, and adopt $\sigma{(\cdot,\cdot)}$ function to get $\beta$ percentile (85th) loss as a selection threshold ${\tau}_t$,
\vspace{-0.1cm}
\begin{equation}
    \tau_t = \sigma(L^{D \cup S}_{select,t},\beta).
\end{equation}
Furthermore, we use model $N_t$ to predict pseudo-labels $\hat{Y}^U_t$ of unlabeled samples, similar to the operation of labeled data mentioned above.
We then use $\omega$ function to map $\hat{Y}^U_t$ to medium exposure to get $\tilde{Y}^{U}_t$ and calculate the loss between 
$\tilde{Y}^{U}_t$ and original medium exposure LDR image $X^{U}_{2,t}$ to get $L^{U}_{select,t} {=} \{L^{U_1}_{select,t},L^{U_2}_{select,t},\dots,L^{U_N}_{select,t}\}$. If the current loss $L^{U_i}_{select,t}$ is greater than ${\tau}_t$,  we consider the pseudo-label to be of poor quality, which has more saturated and ghosted regions. Then we will give a lower weight which tends to decay linearly in the next training iteration.
\vspace{-0.1cm}
\begin{equation}
    L^{U}_{select,t} = ||mask(\omega(\hat{Y}^U_t))-mask(\omega({X}^U_{2,t}))||_1,
\end{equation}
\begin{equation}
    m^U_t = max(L^{U}_{select,t}),
\end{equation}

\begin{equation}
W^{U_i}_{t+1}=
\begin{cases}
1 & \text{ $L^{U_i}_{select,t}$ $\leq$ ${\tau}_t$} \\ 
\frac{m^{U}_{t}-L^{U_i}_{select,t}} {m^{U}_{t}-{\tau}_t} & \text{ $L^{U_i}_{select,t}$ $>$ ${\tau}_t$}
\end{cases}
\end{equation}
where ${X}^U_{2,t}$ is the unlabeled medium exposure image in timestep $t$, $m^U_t$ is the largest selection loss of unlabeled samples in timestep $t$, $W^{U_i}_{t+1}$ is the weight factor of ${U_i}$ smaple in the ${t+1}$ training iteration.

\begin{figure*}[t!]
  \centering
  \includegraphics[width=0.98\linewidth]{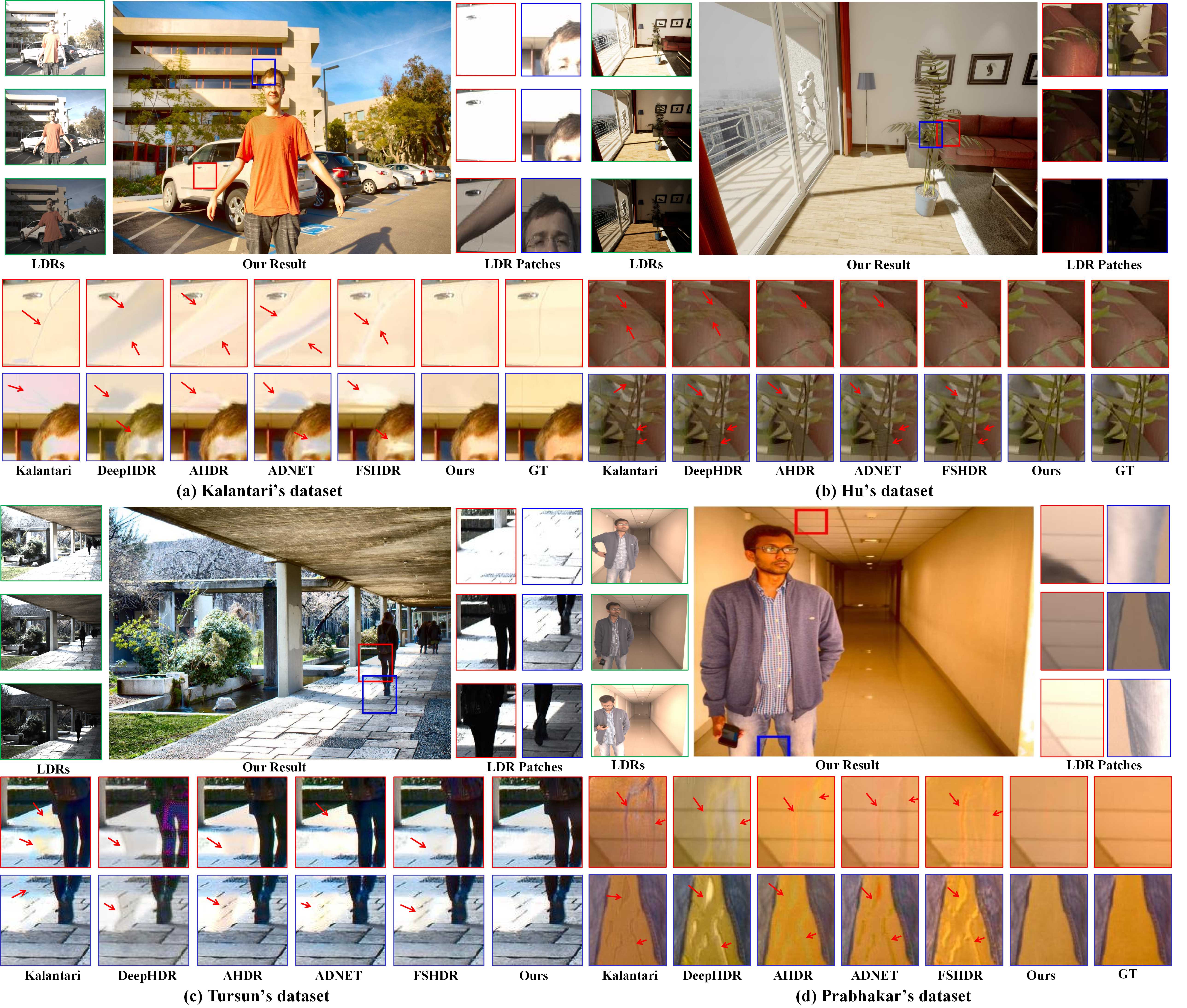}
  \caption{Examples of Kalantari’s \cite{Kalantari2017Deep} and Hu’s \cite{Hu2020Sensor} datasets  (top row) and Tursun’s \cite{Tursun2016An} and Prabhakar's  \cite{prabhakar2019fast} datasets (bottom row). Note that we directly evaluate the methods on Tursun’s and Prabhakar's datasets with the checkpoint trained on Kalantari’s dataset.}
  \vspace{-0.4cm}
  \label{fig:KalanHuTursunSen}
\end{figure*}

\section{Experiments}

\noindent\textbf{Datasets.}
We train all the methods on two public datasets, Kalantari’s \cite{Kalantari2017Deep} and Hu’s dataset \cite{Hu2020Sensor}. Kalantari’s dataset includes 74 training samples and 15 testing samples. Three different LDR images in a sample are captured with exposure biases of $\{$-2, 0, +2$\}$ or $\{$-3, 0, +3$\}$. Hu’s dataset is captured at three exposure levels (\ie,  $\{$-2, 0, +2$\}$). There are 85 training samples and 15 testing samples in Hu’s dataset. We train all comparison methods with the same set of images. Concretely, we randomly choose $K {\in} \{1,5\}$ dynamic labeled samples and $Q{=}5$ static labeled samples for training in all methods. Furthermore, for each $K$, we evaluate all methods for 5 runs denoted as 5-way in Table \ref{table1}. In addition, since FSHDR \cite{prabhakar2021labeled} and our method exploit unlabeled samples, we also use the rest of the dataset samples as unlabeled data $U$. Finally, to verify generalization performance, we evaluate all methods on Tursun’s dataset \cite{Tursun2016An} that does not have ground truth and Prabhakar’s dataset \cite{prabhakar2019fast}.

\noindent\textbf{Evaluation Metrics.}
We calculate five common metrics used for testing, \ie, PSNR-L, PSRN-$\mu$, SSIM-L, SSIM-$\mu$, and HDR-VDP-2 \cite{Mantiuk2011HDR}, where ‘-L’ denotes linear domain, ‘-$\mu$’ denotes tonemapping domain.

\begin{table*}
\footnotesize
\caption{The evaluation results on Kalantari's \cite{Kalantari2017Deep} and Hu's \cite{Hu2020Sensor} datasets. The best and the second best results are highlighted in \textbf{Bold} and \underline{Underline}, respectively.}
\centering
\setlength{\tabcolsep}{1.6mm}
\label{table1}
\begin{tabular}{c|c|c|ccccccc}
\noalign{\smallskip} \hline \noalign{\smallskip}
\textbf{Dataset} & \textbf{Metric} & \textbf{Setting} & \textbf{Kalantari} & \textbf{DeepHDR} & \textbf{AHDRNet} & \textbf{ADNet} & \textbf{FSHDR} & \textbf{Ours} \\
\hline

\specialrule{0em}{1pt}{1pt}
\specialrule{0em}{1pt}{1pt}

\multirow{4}{*}{Kalantari} & PSNR-$l$ & \multirow{2}{*}{5way-5shot} & 39.37$\pm$0.12 &                                    38.25$\pm$0.29 & 40.61$\pm$0.10  &                                               40.78$\pm$0.15  &\underline{41.39}$\pm$0.12                                      &\textbf{41.54}$\pm$0.10 \\
                           & PSNR-$\mu$ & {} & 39.86$\pm$0.19 &                            38.62$\pm$0.27 & 41.05$\pm$0.32  &                              40.93$\pm$0.38  &\underline{41.40}$\pm$0.13                              &\textbf{41.61}$\pm$0.08 \\ \cline{2-9}
                           & PSNR-$l$ & \multirow{2}{*}{5way-1shot} & 36.94$\pm$0.44 &                            36.67$\pm$0.67 & 38.83$\pm$0.39 &                              38.96$\pm$0.35  &\underline{41.04}$\pm$0.11                              &\textbf{41.14}$\pm$0.11 \\
                           & PSNR-$\mu$ & {} & 37.33$\pm$1.21 &                            37.01$\pm$1.68 & 39.15$\pm$1.04  &                              39.08$\pm$1.06  &\underline{41.13}$\pm$0.07                              &\textbf{41.25}$\pm$0.05 \\

\specialrule{0em}{1pt}{1pt}
\hline
\specialrule{0em}{1pt}{1pt}
\specialrule{0em}{1pt}{1pt}

\multirow{4}{*}{Hu} & PSNR-$l$ & \multirow{2}{*}{5way-5shot} & 41.36$\pm$0.25 &                                    40.73$\pm$0.66 & 46.37$\pm$0.76  &                                               46.88$\pm$0.81  &\underline{47.13}$\pm$0.13                                      &\textbf{47.41}$\pm$0.12 \\
                           & PSNR-$\mu$ & {} & 38.95$\pm$0.14 &                            39.92$\pm$0.22 & 43.42$\pm$0.44  &                              43.79$\pm$0.48  &\underline{43.98}$\pm$0.27                              &\textbf{44.24}$\pm$0.17 \\ \cline{2-9}
                           & PSNR-$l$ & \multirow{2}{*}{5way-1shot} & 38.67$\pm$0.43 &                            37.82$\pm$0.86 & 44.64$\pm$0.80 &                              44.75$\pm$0.84  &\underline{44.94}$\pm$0.23                              &\textbf{45.04}$\pm$0.16 \\
                           & PSNR-$\mu$ & {} & 36.83$\pm$0.62 &                            38.49$\pm$1.07 & 42.37$\pm$1.42  &                              42.41$\pm$1.20  &\underline{42.50}$\pm$0.87                              &\textbf{42.55}$\pm$0.44 \\

\noalign{\smallskip} \hline \noalign{\smallskip}
\end{tabular}
\vspace{-0.4cm}
\end{table*}

\begin{table*}
\footnotesize
\caption{Further evaluation results on Kalantari's \cite{Kalantari2017Deep}, Hu's \cite{Hu2020Sensor} and  Prabhakar’s datasets \cite{prabhakar2019fast}. The best and the second best results are highlighted in \textbf{Bold} and \underline{Underline} in each setting, respectively.}
\centering
\setlength{\tabcolsep}{1.6mm}
\label{table2}
\begin{tabular}{cllllll|lllll}
\hline
\multicolumn{6}{c}{\textbf{Kalantari}}  & &  \multicolumn{5}{c}{\textbf{Hu}} \\ \hline
\multicolumn{1}{l}{} &            & PSNR-$l$    & PSNR-$\mu$ & SSIM-$l$ & SSIM-$\mu$ & HV2 & PSNR-$l$ & PSNR-$\mu$ & SSIM-$l$ & SSIM-$\mu$ & HV2 \\ \hline
\multirow{4}{*}{$S_1$}  & Sen        & 38.57     & 40.94   & 0.9711        & 0.9780        & 64.71         & 33.58  & 31.48   & 0.9634       & 0.9531        & 66.39         \\
                     & Hu         & 30.84     & 32.19   & 0.9408       & 0.9632        & 62.05         & 36.94  & 36.56   & 0.9877       & 0.9824        & 67.58         \\
                     & FSHDR      & \underline{40.97}     & \underline{41.11}   & \underline{0.9864}       & \underline{0.9827}        & \underline{67.08}         & \underline{42.15}  & \underline{41.14}   & \underline{0.9904}       & \underline{0.9891}        & \underline{71.35}         \\
                     & Ours (K=0) & \textbf{41.12}     & \textbf{41.20}   & \textbf{0.9866}       & \textbf{0.9868}        & \textbf{67.16}         & \textbf{42.99}  & \textbf{41.30}    & \textbf{0.9912}       & \textbf{0.9903}        & \textbf{72.18}         \\ \hline
\multirow{2}{*}{$S_2$}  & Ours (K=1) & 41.14     & 41.25   & 0.9866       & 0.9869        & 67.20         & 45.04  & 42.55   & 0.9938       & 0.9928        & 73.23         \\
                     & Ours (K=5) & 41.54     & 41.61   & 0.9879       & 0.9880        & 67.33         & 47.41  & 44.24   & 0.9974       & 0.9936        & 74.49         \\ \hline
\multirow{6}{*}{$S_3$}  & Kalantari  & 41.22     & 41.85   & 0.9848       & 0.9872        & 66.23         & 43.76  & 41.60   & 0.9938       & 0.9914        & 72.94         \\
                     & DeepHDR    & 40.91     & 41.64   & 0.9863       & 0.9857        & 67.42         & 41.20  & 41.13    & 0.9941       & 0.9870        & 70.82         \\
                     & AHDRNet    & 41.23     & 41.87   & 0.9868       & \underline{0.9889}        & 67.50         & 49.22  & 45.76   & 0.9980       & \underline{0.9956}        & 75.04         \\
                     & ADNET      & 41.31     & 41.80   & 0.9871       & 0.9883        & 67.57         & \textbf{50.38}  & \underline{46.79}   & \underline{0.9987}       & 0.9948        & \textbf{76.32}         \\
                     & FSHDR      & \textbf{41.79}     & \underline{41.92}   & \underline{0.9876}       & 0.9851        & \underline{67.70}         & 49.56  & 45.90   & 0.9984       & 0.9945        & 75.25         \\
                     & Ours       & \underline{41.68}     & \textbf{41.97}   & \textbf{0.9889}       & \textbf{0.9895}        & \textbf{67.77}         & \underline{50.31}  & \textbf{46.88}   & \textbf{0.9988}       & \textbf{0.9957}        & \underline{76.21}         \\ \hline

\multirow{6}{*}{$S_4$}  & Kalantari  & 25.87     & 21.44   & 0.8610        & 0.9176        & 60.00         & 10.23  & 16.95   & 0.6903       & 0.8346         & 49.10         \\
                     & DeepHDR    & 25.92     & 21.43   & 0.8597       & 0.9170        & 60.02         & \underline{25.48}  & \underline{20.86}   & \underline{0.9215}       & 0.8354        & \underline{66.83}         \\
                     & AHDRNet    & 26.62     & \underline{22.08}   & 0.8737       & \underline{0.9238}        & 58.89         & 11.44  & 17.84   & 0.6732       & 0.8389        & 52.79         \\
                     & ADNET      & 25.76     & 21.39   & 0.8686       & 0.8217        & 60.36         & 10.86     & 18.09   & 0.6915       & \underline{0.8399}        & 49.28         \\
                     & FSHDR      & \textbf{28.03}     & 22.01   & \underline{0.8751}       & 0.9203        & \underline{60.53}         & 12.82     & 19.37   & 0.7442       & 0.8347        & 55.34         \\
                     & Ours       & \underline{27.91}     & \textbf{22.45}   & \textbf{0.8764}       & \textbf{0.9252}        & \textbf{61.02}         & \textbf{30.29}  & \textbf{21.56}   & \textbf{0.9440}       & \textbf{0.8456}        & \textbf{67.07}         \\ \hline    
\multirow{6}{*}{$S_5$}  & Kalantari  & 31.24     & 33.10   & 0.9527       & 0.9593        & 63.99         & 19.82  & 18.63   & 0.7679       & 0.8742        & 59.50         \\
                     & DeepHDR    & 30.75     & 29.01   & 0.9244       & 0.9223        & 63.26         & 19.84  & 18.70    & 0.7698       & 0.8752        & 59.48         \\
                     & AHDRNet    & 31.84     & \underline{33.49}   & \textbf{0.9588}       & 0.9606        & \underline{64.40}         & \textbf{20.80}  & 20.51   & \underline{0.8259}       & 0.9136        & \textbf{59.79}         \\
                     & ADNET      & 31.08     & 33.50   & 0.9536       & \underline{0.9636}        & 63.88         & \underline{20.78}  & \underline{20.80}   & \textbf{0.8268}       & \underline{0.9173}        & 59.71         \\
                     & FSHDR      & \underline{32.70}     & 32.24   & 0.9553       & 0.9465        & 64.37         & 20.23  & 19.71   & 0.7929       & 0.9026        & 59.63         \\
                     & Ours       & \textbf{32.72}     & \textbf{34.49}   & \underline{0.9586}       & \textbf{0.9713}        & \textbf{64.45}         & 20.69  & \textbf{21.96}   & 0.8257       & \textbf{0.9207}        & \underline{59.76}         \\ \hline 

\end{tabular}
\vspace{-0.4cm}
\end{table*}

\noindent\textbf{Implementation Details.}
The window size in MSRSTM is 2$\times$2, 4 $\times$4 and 8 $\times$ 8. In the training stage, we crop the 128 $\times$ 128 patches with stride 64 for the training dataset. We use the Adam optimizer, and set the batch size and learning rate as 4 and 0.0005, receptively. And we set $\beta_{1}$=0.9, $\beta_{2}$=0.999, and $\epsilon$=$1e^{-8}$ in the Adam optimizer. We implement our model using PyTorch with 2 NVIDIA GeForce 3090 GPUs and train for 200 epochs.

\subsection{Comparison with State-of-the-art Methods}
To evaluate our model, we carry out quantitative and qualitative experiments comparing with several state-of-the-art methods, including patch-based classical methods: Sen \cite{Sen2012Robust}, Hu \cite{Hu2013HDR}, and deep learning-based methods: Kalantari \cite{Kalantari2017Deep}, DeepHDR \cite{Wu2018Deep}, AHDRNet \cite{Yan2019attention}, ADNet \cite{Liu2021ADNet}, FSHDR \cite{prabhakar2021labeled}. We use the codes provided by the authors.


\noindent\textbf{Evaluation on Kalantari’s and Hu's Datasets.}
In Figure \ref{fig:KalanHuTursunSen} (a) and (b), we compare our method with other state-of-the-art methods in the 5-shot scenario. 
Due to insufficient labeled samples, large motion, and saturation, most comparing methods suffer from color distortion and ghosting artifacts in these two datasets. Kalantari's method and DeepHDR produce undesirable artifacts and color distortion (see Figure \ref{fig:KalanHuTursunSen} (a)(b)). There are two reasons behind that: misalignment of optical flow and homographies and the lack of labeled data. Although AHDRNet and ADNET are proposed to suppress motion and saturation with attention mechanisms, they cannot reconstruct ghost-free HDR images with few labeled samples. They also produce severe ghosting artifacts (see the red block in Figure \ref{fig:KalanHuTursunSen} (a)(b)). FSHDR exploits unlabeled data to alleviate ghosts under the constraint of a few labeled samples, but it is difficult to handle both ghosting and saturation problems simultaneously. We can see that FSHDR still suffers from ghosting artifacts which leaves an obvious hand artifact in the car (see the red block in Figure \ref{fig:KalanHuTursunSen} (a)). Thanks to the proposed SMAE and sample-quality-based iterative learning strategy, which first address the saturation problems using SMAE and then adaptively sample well-exposed and ghost-free pseudo-labels to handle ghosting problems, we can reconstruct ghost-free HDR images with only a few labeled samples.

The quantitative results under the constraint of few shot scenarios on two dataset are shown in Table \ref{table1}. We report means and 95$\%$ margin of variations for 5 and 1 shot cases across 5 runs. Our method achieves state-of-the-art performance on all metrics of two datasets, while most other methods perform poorly with only a few labeled samples. We show that our proposed method surpasses second-best method by 0.15db and 0.21db in terms of PSNR-$l$ and PSNR-$\mu$ for 5way-5shot setting on Kalantari’s dataset, and it also improves by 0.28db and 0.26db for 5way-5shot setting on Hu’s dataset. For 5way-1shot setting, our method consistently outperforms other approaches on two datasets.

In addition, as shown in Table \ref{table2}, we further compare our method with major HDR deghosting approaches in zero-shot setting $S_1$, few-shot setting $S_2$, and fully supervised setting $S_3$. Note that we use all the dynamic labeled samples without static and unlabeled samples for plain training in setting $S_3$. Our zero-shot approach outperforms other methods in zero-shot setting on two datasets.
It also outperforms some 5-shot and fully supervised methods in most metrics. Finally, our few-shot and fully supervised approaches achieve state-of-the-art performance among two datasets.

\begin{table}
\footnotesize
\caption{Ablation study of 5 shot scenario on Kalantari's dataset.}
\centering
\setlength{\tabcolsep}{1.1mm}
\label{table4}
\begin{tabular}{c|c|ccc}
\noalign{\smallskip} \hline \noalign{\smallskip}
\# & \textbf{Model} & \textbf{PSNR-$l$} & \textbf{PSNR-$\mu$} & \textbf{HDR-VDP-2} \\
\hline
\specialrule{0em}{1pt}{1pt}
\specialrule{0em}{1pt}{1pt}

B1& SSHDR & \textbf{41.54} & \textbf{41.61} & \textbf{67.33}  \\
B2& Stage2Net & 41.31 & 41.43 & 67.21  \\
B3& w/o APSS & 41.49 & 41.45 & 67.29  \\
B4& AHDR$^*$ & 41.48 & 41.51 & 67.30  \\
B5& FSHDR$^*$ & 41.41 & 41.43 & 67.26 \\
B6& Vanilla-AHDR & 40.61 & 41.05 & 66.95 \\
B7& Vanilla-FSHDR & 41.39 & 41.40 & 67.25 \\

\noalign{\smallskip} \hline \noalign{\smallskip}
\end{tabular}
\vspace{-0.5cm}
\end{table}

\noindent\textbf{Evaluation Generalization Across Different Datasets.}
We compare our method against other approaches on Kalantari’s, Hu's, Tursun's, and Prabhakar's datasets to verify generalization performance. We directly evaluate the methods with the checkpoint trained on Kalantari’s dataset and show the qualitative results on Tursun's and Prabhakar's datasets in Figure \ref{fig:KalanHuTursunSen} (c)(d). More results are included in the Appendix. In Figure \ref{fig:KalanHuTursunSen} (c), since the lady’s motion is large, all the comparison methods cannot remove the ghosting artifacts. In Figure \ref{fig:KalanHuTursunSen} (d), the comparison methods have obvious color distortion and ghosting artifacts on the floor and in the ceil. It shows that other methods have poor generalization performance across different datasets. All these methods address both the saturation and ghosting problems simultaneously. They cannot learn a robust representation to reconstruct a high-quality HDR image. Thanks to our SMAE and sample-quality-based iterative learning strategy, we can learn a robust representation to recover saturated regions and remove ghosting artifacts.

In Table \ref{table2}, setting $S_4$ denotes that we utilize the checkpoint trained on Kalantari’s or Hu's dataset under 5 shot scenario to evaluate on Hu's or Kalantari’s dataset reversely. Setting $S_5$ represents that we train on Kalantari’s or Hu's dataset under 5 shot scenario and evaluate on Prabhakar's dataset. Our method achieves better numerical performance in terms of PSNR-$l$ and PSNR-$\mu$. It demonstrates that our method generalizes well across different datasets.






\subsection{Ablation Studies}
We conduct ablation studies on Kalantari’s dataset under the condition of 5 shot scenario across 5 runs and analyze the importance of each component. We use the following variants of our whole SSHDR model: 1) {SSHDR}: The full model of SSHDR network trained with two entire stages. 2) {Stage2Net}: The model only trained in the second stage without SMAE pre-training. 3) {w/o APSS}: The model trained with two stages without using sample-quality-based pseudo-labels selection strategy. 4) {AHDR$^*$}: The AHDR model is trained with our proposed two stages strategy. 5) {FSHDR$^*$}: Our model is trained with the FSHDR strategy. 6) {Vanilla-AHDR}: The vanilla AHDR model trained in 5 shot scenario. 7) {Vanilla-FSHDR}: The vanilla FSHDR model trained with 5 labeled samples.

\begin{figure}[!t]
    \centering
    \includegraphics[width=0.9\linewidth]{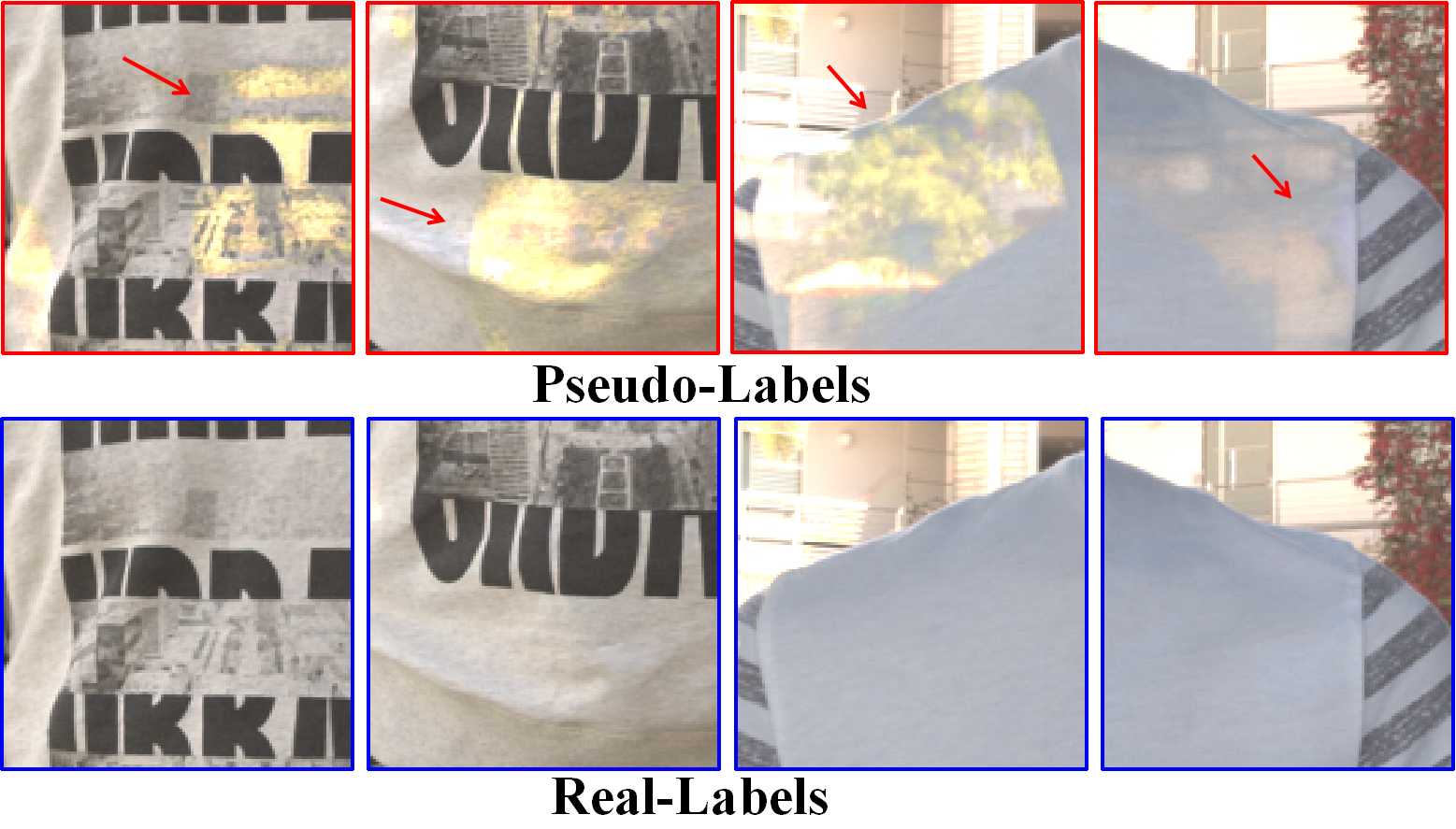}
    \caption{Visual results of poor pseudo-labels.}
    \vspace{-0.3cm}
    \label{fig:ablation}
\end{figure}


\noindent\textbf{SMAE Pre-training.} As shown in Table \ref{table4}, the performance of Stage2Net is significantly decreased compared with SSHDR. Since the SMAE learns a robust representation to generate content of saturated regions, it helps to improve the saturated regions. In a word, it demonstrates that the SMAE pre-training stage is an effective mechanism. 

\noindent\textbf{Pseudo-labels Selection Strategy.} Since the sample-quality-based pseudo-labels selection strategy can exclude saturated and ghosted samples (see Figure \ref{fig:ablation}), the model can be guided in a correct optimization direction which is effective for ghost removal. When we remove the pseudo-labels selection strategy, the performance of the model without APSS is dropped. 

\noindent\textbf{Two Stages Strategy.}
In Table \ref{table4}, we report the performance of AHDR$^*$. It achieves a significant increment compared with the vanilla AHDR model, which demonstrates the effectiveness of the overall two stages strategy. 

\noindent\textbf{Proposed Model Architecture.}
When we replace our two stages strategy with the FSHDR strategy, the numerical results increase compared with FSHDR. It shows that our proposed model architecture is also sound.

\vspace{0.2cm}
\section{Conclusion}
\vspace{0.2cm}

We propose a novel semi-supervised deghosting method for few-shot HDR problem via two stages of completing saturation and deghosting. In the first stage, a Saturated Mask AutoEncoder is proposed to learn a robust representation and reconstruct a non-saturated HDR image with a self-supervised mechanism. In the second stage, we propose an adaptive pseudo-label selection strategy to avoid the effects of mislabeled samples. Finally, our approach shows superiority over the existing state-of-the-art methods.



\clearpage
{\small
\bibliographystyle{ieee_fullname}
\bibliography{egbib}
}

\end{document}